%% file: main.tex
\documentclass[twoside,11pt]{article}

\usepackage{jmlr2e}
\usepackage{amsmath}
\usepackage{amsfonts}
\usepackage{comment}
\usepackage{caption}
\usepackage{subcaption}
\usepackage{multirow}
\usepackage{hyperref}
\usepackage{booktabs}
\usepackage[ruled,vlined]{algorithm2e}

\jmlrheading{1}{}{}{}{}{}

\ShortHeadings{Transferable Candidate Proposal with Bounded Uncertainty}{Go and Kim}
\firstpageno{1}

\begin{document}

\title{Transferable Candidate Proposal with Bounded Uncertainty}

\author{\name Kyeongryeol Go \email krgo@superb-ai.com \\
        \name Kye-Hyeon Kim \email khkim@superb-ai.com \\
        \addr Superb AI\\
        Seoul, South Korea}


\maketitle

\begin{abstract}
From an empirical perspective, the subset chosen through active learning cannot guarantee an advantage over random sampling when transferred to another model. While it underscores the significance of verifying transferability, experimental design from previous works often neglected that the informativeness of a data subset can change over model configurations. To tackle this issue, we introduce a new experimental design, coined as Candidate Proposal, to find transferable data candidates from which active learning algorithms choose the informative subset. Correspondingly, a data selection algorithm is proposed, namely Transferable candidate proposal with Bounded Uncertainty (TBU), which constrains the pool of transferable data candidates by filtering out the presumably redundant data points based on uncertainty estimation. We verified the validity of TBU in image classification benchmarks, including CIFAR-10/100 and SVHN. When transferred to different model configurations, TBU consistency improves performance in existing active learning algorithms. Our code is available at \url{https://github.com/gokyeongryeol/TBU}.
\end{abstract}

\begin{keywords}
  Active Learning, Transferability, Uncertainty Estimation
\end{keywords}

\input{Main_Body/1_Introduction}
\input{Main_Body/2_Related_Works}
\input{Main_Body/3_Methodology}
\input{Main_Body/4_Experiments}
\input{Main_Body/5_Conclusion}

\newpage



\vskip 0.2in
\bibliography{reference}

\newpage

\appendix

\input{Appendix/A_Background}
\input{Appendix/B_Pipeline}
\input{Appendix/C_Laplace}
\input{Appendix/D_FreeMatch}
\input{Appendix/E_Settings}
\input{Appendix/F_Results}


\end{document}

%% file: Main_Body/1_Introduction.tex
\section{Introduction}
\label{sec:introduction}

\begin{figure}[t]
    \centering
    \includegraphics[width=0.88\textwidth]{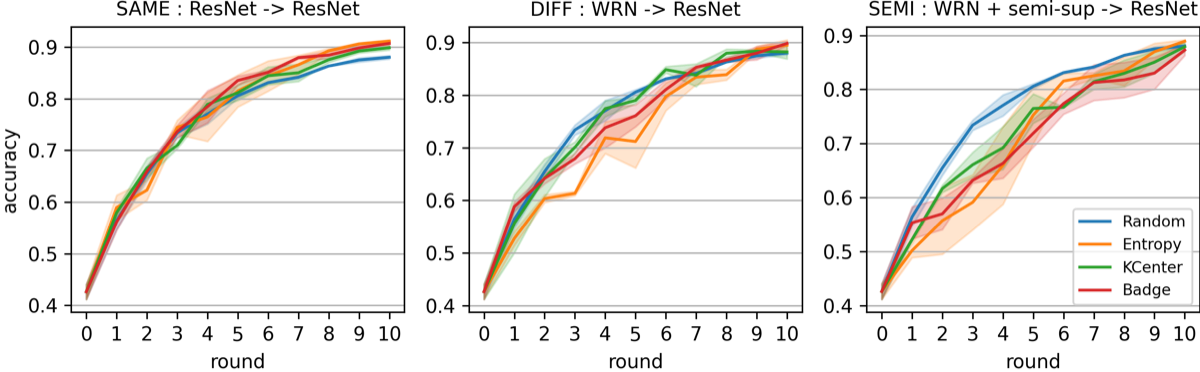}
    \caption{Per-round accuracy of distinct active learning algorithms in CIFAR-10. Regarding the dominance relationship and the resulting performance improvement, active sets achieve data efficiency over random sampling only when the proxy-target pair coincides (left). Transferability is not guaranteed when differing in architecture (middle) or learning algorithm (right). For a detailed explanation of SAME, DIFF, and SEMI, see \autoref{sec: experiment}.}
    \label{fig: criticism}
\end{figure}

Recent empirical observations show that the subset chosen through active learning, referred to as the active set, does not maintain its utility across different model architectures and experimental settings \citep{lowell2018transferable, munjal2022towards, ji2023randomness}. Our pilot study, as illustrated in \autoref{fig: criticism}, also reveals that when there is a mismatch in architectures or learning algorithms, a subset chosen by a proxy model offers no advantage over random sampling when fed to target models. 

\cite{ji2023randomness} underlined the necessity to verify the transferability of the active set and encouraged the evaluation of the target model performance on selected subsets by the proxy model. However, such experimental design expects that there are universally informative data subsets regardless of model configurations, which we argue to be unrealistic. To illustrate the point, consider that many machine learning algorithms operate under the assumption that the effectiveness of individual data points to model training is not solely dependent on the data set itself. Instead, it encompasses numerous factors such as network architecture, learning algorithm, and attained performance. For example, curriculum learning \citep{soviany2022curriculum} and the adaptive data selection framework \citep{mirzasoleiman2020coresets} propose that orchestrating the data loading process can yield tangible benefits. Moreover, in semi-supervised learning, \cite{sohn2020fixmatch, wang2022freematch} selectively choose a subset of unlabeled data to facilitate regularization. These suggest that the informativeness of data subsets evolves dynamically as the learning proceeds.

In this respect, we introduce a new experimental design called Candidate Proposal, which can verify the transferability depending on model configurations. Compared to the previous experimental design, the limited awareness of the proxy model regarding target models is compensated by refining the role of the proxy model while expanding the role of target models. Specifically, the proxy model only contributes to constraining a pool of potentially transferable data candidates within unlabeled data. Then, among those candidates, target models choose the informative subset of a fixed budget using active learning algorithms. Resorting to the training signal from the proxy model, the size of transferable data candidates can exceed the budget, and whether each element of candidates is informative depends on the target model configuration.

Besides the active learning algorithms for subset selection, Candidate Proposal additionally requires a candidate selection algorithm. We propose \textbf{Transferable candidate proposal with Bounded Uncertainty (TBU)} that filters out non-transferable data candidates using uncertainty estimation. Instances of \underline{l}ow \underline{e}pistemic uncertainty (in brief, LE instances) offer minimal information gain \citep{zhou2022survey}, whereas instances of \underline{h}igh \underline{a}leatoric uncertainty (in brief, HA instances) are susceptible to adversarial perturbations \citep{smith2018understanding}. TBU chooses data candidates from sources other than the LE and HA instances, preventing the inclusion of redundant data that provides marginal enhancements to model performance. The validity of TBU is demonstrated in image classification benchmarks by complementary effects on existing active learning algorithms.

%% file: Main_Body/2_Related_Works.tex
\section{Related Works}
\label{sec:related_works}

Conventional approaches, represented by Entropy \citep{gal2017deep} and CoreSet \citep{sener2017active}, selected subsets based on either uncertainty or diversity. Despite their longstanding relevance, these methods remain competitive baselines in contemporary research. To achieve better-calibrated uncertainty estimations, newer techniques have emerged. Monte-Carlo Dropout \citep{gal2016dropout} and Deep Ensemble \citep{lakshminarayanan2017simple}, for instance, incorporate architectural modifications to enhance uncertainty estimation \citep{beluch2018power}. Additionally, approaches like \cite{houlsby2011bayesian, beluch2018power, kirsch2019batchbald} have quantified information gain through predictive entropy. 
Also, gradient information has been leveraged, combining uncertainty and diversity in a hybrid manner. Notably, Badge \citep{ash2019deep} used the K-means++ clustering in the gradient embedding space, and \cite{liu2021influence, wang2022deep} quantified the expected change in model performance using the influence function. Commonly, most of the previous studies have been dedicated to better exploiting the captured training signal. Please refer to \hyperref[app: background]{Appendix A} for a more detailed background on the acquisition functions.


The most similar work is the data valuation framework \citep{ghorbani2019data, kwon2021beta, wang2023data}, which compares the relative contribution of each training data point to clarify both the most effective and defective data points regardless of model configurations. Yet, due to expensive computational costs, their practical usage is limited to low-dimensional data. To the best of our knowledge, this is the first work on verifying the transferability of the Candidate Proposal in active learning.

%% file: Main_Body/3_Methodology.tex
\section{Methodology}
\label{sec: methodology}

\begin{figure}
    \centering
    \includegraphics[width=0.88\textwidth]{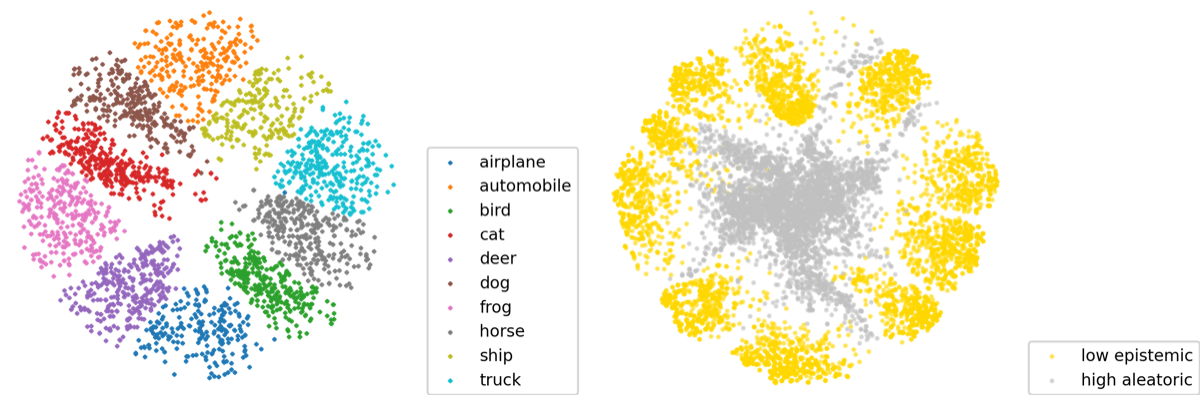}
    \caption{T-SNE embedding of the labeled (left) and the instances of the unlabeled filtered by TBU (right). Different colors indicate different class labels for the labeled data embedding, while the LE and HA instances in the unlabeled data embedding are denoted with gold and silver, respectively.}
    \label{fig: embedding}
\end{figure}

This section explains TBU, the candidate selection algorithm for Candidate Proposal. TBU filters out LE and HA instances using deterministic uncertainty methods and semi-supervised learning. As illustrated in \autoref{fig: embedding}, LE instances tend to have far margins from the decision boundaries while HA instances have a small margin from the decision boundary. Please refer to \hyperref[app: pipeline]{Appendix B} for the overall pipeline.

\subsection{Quantifying Epistemic Uncertainty}

In estimating epistemic uncertainty, we resort to the deterministic uncertainty methods \citep{liu2020simple, van2021feature, mukhoti2023deep} which has been regarded as a valid substitute for Bayesian Neural Networks (BNN). Among various choices, we applied the last-layer Laplace approximation \citep{kristiadi2020being} for its simplicity and scalability. 

The last-layer Laplace approximation \citep{kristiadi2020being, daxberger2021laplace} falls into the category that replaces the final soft-max layer with a distance-sensitive function. Equipping the parameter $w=[w_1,...,w_C]$ of the linear classifier with prior $p(w_c)=\mathcal{N}(w_c;0, I)$ and posterior $p(w_c|D)=\mathcal{N}(w_c;\mu_c, \Sigma_c)$, the proxy network is now composed of a fixed feature extractor $\phi$ and a bayesian linear classifier. The parameter of posterior distribution can be estimated via Laplace approximation. Specifically, $\mu=[\mu_1, ..., \mu_C]$ is the Maximum A Posteriori (MAP) estimate, and $\Sigma=[\Sigma_1,...,\Sigma_C]$ is the inverse of the precision matrix that is computed as the Hessian of negative log posterior at $\mu$.
\begin{align*}
    \mu &= \arg\max_{w} \log {p(D|w) p(w)} \\
    \Sigma &= (-\nabla_w^2 \log {p(D|w)p(w)}\vert_{w=\mu})^{-1}
\end{align*}
Note that $D$ denotes the train data and $p(D|w)=\prod_{(x,y)\in D} p(y|\phi(x),w)$ implies the likelihood. Then, the model prediction turns to marginalization over the posterior such that  
\begin{equation*}
    p(y|x,D) = \int p \left(y|\phi(x), w \right) p(w|D) dw = \int \sigma \left( \phi(x)^T w \right)  p(w|D) dw 
\end{equation*}
where $\sigma$ implies the soft-max activation function. Please refer to \hyperref[app: laplace]{Appendix C} for details on the approximation for the inference process.

We regard the unlabeled data instances as having low epistemic uncertainty if their predictive entropy is small enough. Specifically, for each predicted class label, the unlabeled data instances whose predictive entropy is lower than $q$-th percentile of that of the labeled data set are categorized as LE instances.

\subsection{Identifying Instances of High Aleatoric Uncertainty}

Instances of high aleatoric uncertainty can be inspected more straightforwardly. Unlike previous approaches that explicitly incorporate complex noise distribution using parameter-intensive modules \citep{fortuin2021deep, collier2021correlated, collier2023massively}, we rely on semi-supervised learning, harnessing the unlabeled data to identify HA instances effectively. In short, we employ the class-wise confidence threshold proposed in FreeMatch \citep{wang2022freematch} to set an upper limit on the predictive confidence. 

Free-match \citep{wang2022freematch} suggested iteratively updating the class-wise confidence threshold to lower bound the beneficial subset of the unlabeled data set. For every iteration $t$, global threshold $g^{(t)} \in \mathbb{R}$ and the local threshold $l^{(t)}=[l_{1}^{(t)}, ..., l_{C}^{(t)}] \in \mathbb{R}^{C}$ are updated using the model predictive confidence $[p_1, ..., p_C] \in \mathbb{R}^{C}$ as follows:
\begin{align*}
    g^{(t)} & = \eta \cdot g^{(t-1)} + (1-\eta) \cdot \frac{1}{\vert S \vert} \sum_{s=1}^{\vert S \vert} max(p_1,...,p_C) \\
    l_c^{(t)} & = \eta \cdot l_c^{(t-1)} + (1-\eta) \cdot \frac{1}{\vert S \vert} \sum_{s=1}^{\vert S \vert} p_c \quad \forall c \in [C]
\end{align*}
Note that $g^{(0)}$ and every element of $l^{(0)}$ are initialized to $\frac{1}{C}$, and $\eta$ denotes the coefficient of the exponential moving average. The local threshold then modulates the global threshold to compute the class-wise confidence threshold.
\begin{equation*}
    \tau_c^{(t)} = g^{(t)} \cdot l_c^{(t)} / max ( l_1^{(t)}, ..., l_C^{(t)} )
\end{equation*}

We used this threshold to upper bound the predictive confidence of HA instances in the unlabeled data set since those are uncertain enough to be ignored throughout model training. Practically, considering that the model prediction can drastically change across epochs \citep{toneva2018empirical}, only the unlabeled instances that consistently show low confidence for five times of periodic evaluation at every 12.5$\%$ of total epochs are regarded to have high aleatoric uncertainty. Please refer to \hyperref[app: freematch]{Appendix D} detailing the rationale in class-wise confidence threshold.

%% file: Main_Body/4_Experiments.tex
\section{Experiment}
\label{sec: experiment}

\begin{table}[t]
    \centering
    \small
    \begin{tabular}{ |c|c|c|c|c|c|c| }
        \hline
        & & 1st & 2nd & 3rd & 4th \\
        \hline
        \multirow{6}{*}{Entropy} & SAME & \underline{0.589 $\pm$ 0.024} & 0.623 $\pm$ 0.019 & 0.745 $\pm$ 0.014 & 0.793 $\pm$ 0.026 \\
        & DIFF & 0.527 $\pm$ 0.013 & 0.603 $\pm$ 0.007 & 0.613 $\pm$ 0.004 & 0.710 $\pm$ 0.027 \\
        & SEMI & 0.502 $\pm$ 0.013 & 0.557 $\pm$ 0.062 & 0.591 $\pm$ 0.051 & 0.648 $\pm$ 0.034 \\
        & TBU($q$=10) & 0.573 $\pm$ 0.021 & \underline{0.682 $\pm$ 0.039} & \underline{0.766 $\pm$ 0.018} & \underline{0.807 $\pm$ 0.015} \\
        & TBU($q$=20) & \textbf{0.617 $\pm$ 0.006} & 0.653 $\pm$ 0.065 & \textbf{0.768 $\pm$ 0.022} & \textbf{0.814 $\pm$ 0.010} \\
        & TBU($q$=50) & 0.546 $\pm$ 0.032 & \textbf{0.701 $\pm$ 0.018} & 0.759 $\pm$ 0.014 & 0.804 $\pm$ 0.010 \\
        \hline
        \multirow{6}{*}{CoreSet} & SAME & 0.579 $\pm$ 0.008 & 0.666 $\pm$ 0.019 & 0.710 $\pm$ 0.006 & \underline{0.786 $\pm$ 0.018} \\
        & DIFF & 0.556 $\pm$ 0.056 & 0.643 $\pm$ 0.036 & 0.701 $\pm$ 0.011 & 0.760 $\pm$ 0.031 \\
        & SEMI & 0.522 $\pm$ 0.003 & 0.617 $\pm$ 0.007 & 0.661 $\pm$ 0.023 & 0.707 $\pm$ 0.028 \\
        & TBU($q$=10) & 0.568 $\pm$ 0.030 & \textbf{0.675 $\pm$ 0.004} & \textbf{0.743 $\pm$ 0.028} & \textbf{0.796 $\pm$ 0.013} \\
        & TBU($q$=20) & \textbf{0.608 $\pm$ 0.005} & \underline{0.674 $\pm$ 0.005} & 0.733 $\pm$ 0.024 & 0.784 $\pm$ 0.010 \\
        & TBU($q$=50) & \underline{0.586 $\pm$ 0.024} & 0.654 $\pm$ 0.024 & \underline{0.740 $\pm$ 0.008} & 0.768 $\pm$ 0.011 \\
        \hline
        \multirow{6}{*}{Badge} & SAME & 0.561 $\pm$ 0.016 & 0.659 $\pm$ 0.009 & 0.736 $\pm$ 0.012 & 0.784 $\pm$ 0.022 \\
        & DIFF & \underline{0.588 $\pm$ 0.009} & 0.642 $\pm$ 0.010 & 0.679 $\pm$ 0.010 & 0.766 $\pm$ 0.025 \\
        & SEMI & 0.553 $\pm$ 0.030 & 0.570 $\pm$ 0.029 & 0.632 $\pm$ 0.005 & 0.708 $\pm$ 0.020 \\
        & TBU($q$=10) & \textbf{0.601 $\pm$ 0.030} & \underline{0.680 $\pm$ 0.028} & \underline{0.744 $\pm$ 0.009} & \textbf{0.819 $\pm$ 0.013} \\
        & TBU($q$=20) & 0.584 $\pm$ 0.036 & \textbf{0.696 $\pm$ 0.010} & \textbf{0.754 $\pm$ 0.024} & \underline{0.797 $\pm$ 0.013} \\
        & TBU($q$=50) & 0.566 $\pm$ 0.018 & 0.672 $\pm$ 0.014 & 0.743 $\pm$ 0.021 & 0.780 $\pm$ 0.037 \\
        \hline
    \end{tabular}
    \caption{Per-round accuracy of the target model with ablation of the percentile $q$ in TBU. The first and the second highest values are marked in bold and underlined, respectively.}
    \label{tab: sensitivity-cifar10}
\end{table}

We conducted experiments on several image classification benchmarks in active learning, including CIFAR-10/100 and SVHN. In every round, proxy and target models are trained from scratch with a cumulative collection of the selected subsets from the previous rounds. In between rounds, the proxy model conducts the candidate proposal by filtering out the redundant instances using TBU, and the target model selects the subset of a fixed budget within those candidates. To verify the efficacy of TBU, we considered three baselines, namely SAME, DIFF, and SEMI. These do not use the candidate proposal but select a fixed-sized subset either from the proxy or the target.\footnote{Unless otherwise specified, all the baseline models are trained using supervised learning.}
\begin{itemize}
    \item SAME: The proxy and the target coincide. In other words, the target selects the subset from the unconstrained pool of unlabeled data.
    \item DIFF: The proxy and the target have different architectures, and the proxy selects a fixed-sized subset on behalf of the target.
    \item SEMI: Keeping all other conditions identical to DIFF, the proxy is trained by semi-supervised learning.
\end{itemize}

For generality, active learning algorithms of distinct categories (uncertain, diverse, hybrid), represented by Entropy, CoreSet, and Badge, are considered. For reproducibility, we made our codes available. Please refer to \hyperref[app: settings]{Appendix E} for experimental settings.

In \autoref{tab: sensitivity-cifar10}, we reported the per-round accuracy of the target model in CIFAR-10, revealing three intriguing outcomes. First, SEMI appears to fall short when compared to DIFF. This outcome might strike one as counter-intuitive, given that more advanced learning techniques, such as semi-supervised learning, typically lead to a deeper understanding of the underlying data distribution. We posit this result as empirical evidence illustrating that the informativeness of a data point is not solely dictated by the data set itself; instead, it can vary depending on the specific model configurations employed. Second, TBU has the potential to surpass SAME in performance, irrespective of the active learning algorithms. This suggests that, without altering the acquisition function, the effectiveness of existing active learning algorithms can be enhanced by constraining the search space of target models through the proxy model using TBU. Third, TBU is robust to the choice of the percentile $q$ unless it is set too high. Throughout the experiments, we fix $q$ to 10.

Beyond the findings from per-round accuracy in CIFAR-10, we have investigated TBU through experiments in various aspects. To begin with, for generality, we demonstrated that TBU could achieve higher per-round accuracy over baselines even when varying data sets and target model architectures. Furthermore, the trained model with TBU is shown to be robust to input perturbation when evaluated in CIFAR-10/100-C. Next, for qualitative analysis, instances of low epistemic uncertainty exhibit semantic similarity in their imagery, whereas instances of high aleatoric uncertainty incorporate fragmented object components or exhibit a distinct foreground/background color palette. Lastly, TBU could automatically schedule the difficulty of subsets along selection rounds. Due to space constraints, we report these experimental results in \hyperref[app: results]{Appendix F}.

%% file: Main_Body/5_Conclusion.tex
\section{Conclusion}
\label{sec: conclusion}

Although active learning has a rich historical background and enjoys consensus among researchers regarding its importance, the application of active learning remains a challenge due to empirical criticisms. We aim to expand upon the previous experimental design by the Candidate Proposal and propose a candidate selection algorithm, namely Transferable candidate proposal with Bounded Uncertainty (TBU). Throughout the experiments, it is demonstrated that TBU can seamlessly integrate with existing active learning algorithms, offering consistent performance improvement.

%% file: Appendix/A_Background.tex
\section*{Appendix A. Background on Acquisition Functions}
\label{app: background}

Along with transfer learning, semi-supervised learning, and data augmentation, active learning aims to train machine learning models with limited labeled data effectively. Concretely, let $L=\{ (x_l,y_l) \}$ and $U_x=\{x_u\}$ denote the labeled and unlabeled data set. The model is trained from scratch with $L$ using supervised learning or $L$ and $U_x$ using semi-supervised learning at each iteration. Then, a model-dependent acquisition function selects a subset of size K from $U_x$, labeled by an oracle to be added to $L$. 

Over the past few years, deep neural networks have been highly regarded in active learning for their flexibility as function approximators and broad applicability in various data analysis tasks, such as dimensionality reduction and uncertainty estimation \citep{liu2017survey}. Therefore, active learning research has focused on developing improved acquisition functions using deep neural networks, namely deep active learning \citep{ren2021survey}.

The acquisition function, also known as the utility, is crucial in quantifying the informativeness for selecting uncertain and diverse subsets. In essence, it is a set function, and its purpose is to identify the subset that is most likely to yield maximal performance improvements within a fixed budget. However, traversing and comparing all possible combinations of an unlabeled data set is not computationally scalable. Consequently, many approaches concentrate on evaluating each data instance independently by considering subsets of size one and selecting the top-K instances based on their scores \citep{gal2017deep, yoo2019learning, liu2021influence, wang2022deep}. Alternatively, another approach uses an acquisition function that exhibits the properties of monotonicity and sub-modularity, which transforms the problem into sub-modular optimization \citep{sener2017active, kirsch2019batchbald, mirzasoleiman2020coresets}. By employing a greedy selection strategy, these methods achieve a solution with only a small approximation error compared to the optimal solution \citep{nemhauser1978analysis}.

Following are the active learning algorithms we have considered throughout the experiments, which are representative of distinct categories (uncertainty, diversity, hybrid).
\begin{itemize}
    \item Entropy \citep{gal2017deep}: top-K samples of predictive entropy
    \item CoreSet \citep{sener2017active}: K greedy selection based on the maximum distances between the feature embeddings of the labeled and unlabeled data sets
    \item Badge \citep{ash2019deep}: K-means++ clustering results in gradient embedding space with respect to the weight parameter of the last linear layer
\end{itemize}

%% file: Appendix/B_Pipeline.tex
\section*{Appendix B. Pipeline of Candidate Proposal with TBU}
\label{app: pipeline}

\begin{figure}[h]
    \centering
    \includegraphics[width=\textwidth]{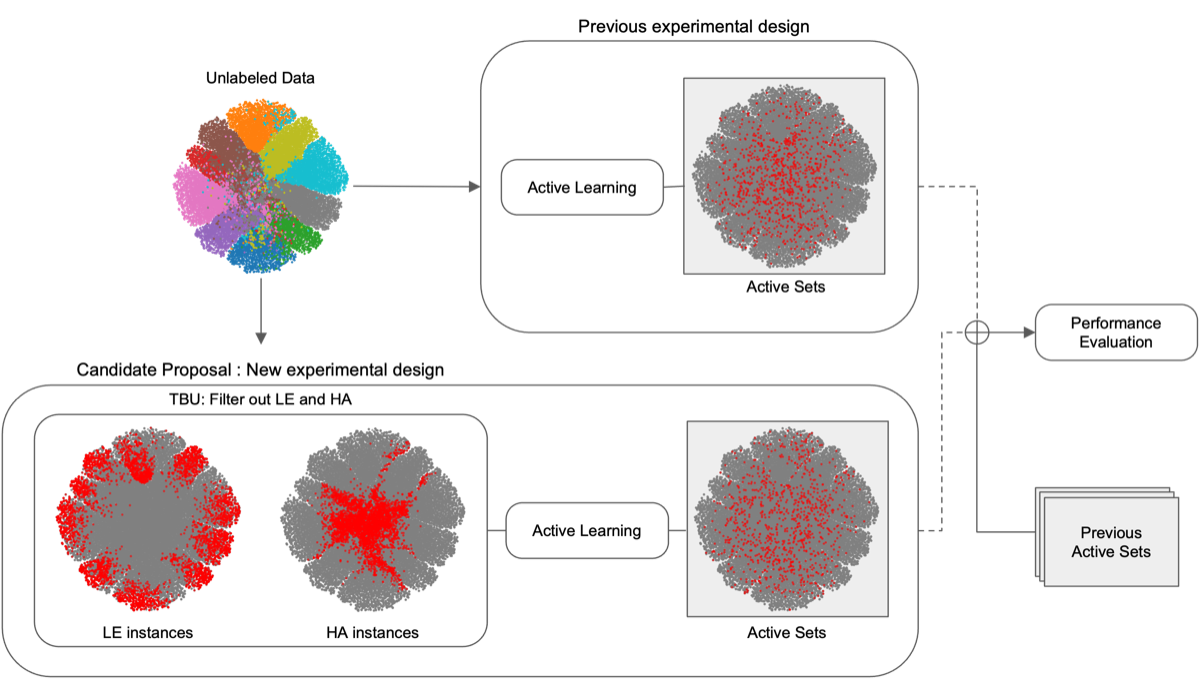}
    \caption{An overall pipeline of Candidate Proposal with TBU in comparison to the previous experimental design. TSNE embeddings of unlabeled data are attached along the pipeline. The unlabeled data in the top-left corner is colored by the ground-truth class labels. For the rest, the data points corresponding to the text description written below are colored red while the remaining are colored gray.}
    \label{fig: pipeline}
\end{figure}

In \autoref{fig: pipeline}, we compare the previous experimental design and our new experimental design, Candidate Proposal. Given an unlabeled data set, previous experimental designs apply active learning on training signals from the proxy model to choose an informative subset for target models. In Candidate Proposal, transferable data candidates are first determined by TBU that filters out LE and HA instances, then an informative subset is chosen from the candidates by target models. Please refer to \autoref{alg: TBU} for the step-by-step description.

Now, we describe how TBU differs from the three baselines that appeared in \autoref{sec: experiment}, SAME, DIFF, and SEMI, which correspond to the previous experimental design. While DIFF and SEMI assume different model configuration for the proxy and target models, SAME use the same configuration. Therefore, SAME is the only baseline where the target model selects the subset by itself. Likewise, TBU also allows the target model to directly search the informative subset. However, unlike SAME, the search space is shrunk from the whole unlabeled data set by filtering out presumably redundant data points using the proxy model of different configurations.

\begin{algorithm}[!ht]
    \SetKwInput{KwData}{Input}
    \SetKwInput{KwResult}{Output}
    \SetAlgoLined
    $\mathbf{Dataset:}$ labeled, unlabeled, validation data set: $L, U_x, V$ \\
    $\mathbf{Model:}$ proxy, target models: $f_p, f_t$ \\

    \KwData{acquisition function: $\mathcal{A}$, scaling factor: $\lambda$, percentile: $q$, budget size: $K$}
    \KwResult{acquisition batch: $S$}

    Randomly initialize the parameters of proxy and target models \\

    \While{\textit{not converged}}{
        Update parameters of the proxy model $f_p$ with $L$ and $U_x$ \\
        Following \autoref{eq: global-local}, update global and local threshold
    }
    
    Following \autoref{eq: scalability}, compute the logit variance $\hat{\Sigma}$ with $L$ \\
    Following \autoref{eq: prediction}, use L-BFGS optimizer for $\lambda$ to minimize NLL in $V$ \\

    Inspect LE instances $S_{LE}$ of lower predictive entropy than the $q$-th percentile in $L$ \\
    Inspect HA instances $S_{HA}$ of lower predictive confidence than \autoref{eq: threshold} \\

    \While{\textit{not converged}}{
        Update parameters of the target model $f_t$ with $L$
    }
    $S \leftarrow \emptyset$, $C \leftarrow U_x - (S_{LE} \cup S_{HA})$ \\
    \For{$k=1,...,K$}{
        $s = \arg\max_{c}{\mathcal{A}(f_t, x_c)}$ for $x_c \in C$\\
        $S \leftarrow S \cup \{x_s\}, C \leftarrow C - \{x_s\}$
    }

    \caption{Transferable Candidate Proposal with Bounded Uncertainty (TBU)}
    \label{alg: TBU}
\end{algorithm}

%% file: Appendix/C_Laplace.tex
\section*{Appendix C. Last-layer Laplace Approximation}
\label{app: laplace}

The last-layer Laplace approximation \citep{kristiadi2020being, daxberger2021laplace} falls into the category that replaces the final soft-max layer with a distance-sensitive function. Equipping the parameter $w=[w_1,...,w_C] \in \mathbb{R}^{F \times C}$ of the linear classifier with prior $p(w_c)=\mathcal{N}(w_c;0, I)$ and posterior $p(w_c|D)=\mathcal{N}(w_c;\mu_c, \Sigma_c)$, the proxy network is now composed of a fixed feature extractor $\phi$ and a bayesian linear classifier. Then, the model prediction turns to marginalization over the posterior such that  
\begin{equation}
    \label{eq: marginalize}
    p(y|x,D) = \int p \left(y|\phi(x), w \right) p(w|D) dw = \int \sigma \left( \phi(x)^T w \right)  p(w|D) dw 
\end{equation}
where $D$ is the train data and $\sigma$ implies the soft-max activation function. The parameter of posterior distribution can be estimated via Laplace approximation. Specifically, $\mu=[\mu_1, ..., \mu_C]$ is the Maximum A Posteriori (MAP) estimate, and $\Sigma=[\Sigma_1,...,\Sigma_C]$ is the inverse of the precision matrix that is computed as the Hessian of negative log posterior at $\mu$.
\begin{align*}
    \label{eq: laplace}
    \mu &= \arg\max_{w} \log {p(D|w) p(w)} \\
    \Sigma &= (-\nabla_w^2 \log {p(D|w)p(w)}\vert_{w=\mu})^{-1}
\end{align*}
Note that $p(D|w)=\prod_{(x,y)\in D} p(y|\phi(x),w)$ denotes the likelihood. $\mu$ can be obtained by standard neural network training with L2 regularization, and $\Sigma$ is computed after model training as post-processing, which has a closed-form expression that does not require a backward pass for computing the Hessian. Considering the scalability in a dataset of too many classes, the shared covariance $\hat{\Sigma}$ is used throughout the experiments.
\begin{equation}
    \label{eq: scalability}
    \hat{\Sigma}= \left( \sum_{(x,y)\in D} p^* ( 1-p^* ) \phi(x) \phi(x)^T \right)^{-1}
\end{equation}
This is proven to be an upper bound for all class-specific covariances $\Sigma_c$, where $p^*=\max{(p_1,...,p_C)}$ denotes the predictive confidence \citep{liu2023simple}. The covariance is estimated only once before the inference phase, which is computationally affordable compared to other efficient ensemble methods \citep{wen2020batchensemble, dusenberry2020efficient}.

Mean-field approximation for Gaussian-Softmax integral \citep{lu2020mean} can play a role as a decent replacement of Monte-Carlo averaging for \autoref{eq: marginalize}
\begin{equation}
    \label{eq: prediction}
    p(y|x,D) = \int \sigma(l) \: \mathcal{N} \left( l; \phi(x)^T \mu, v(x) \right) dl \approx \sigma \left( \frac{\phi(x)^T \mu}{\sqrt{1+\lambda \cdot v(x)}} \right)
\end{equation}
where $\lambda$ is a scaling hyper-parameter. Note that this enables the inference by a single forward pass. Regarding the symbols, $v(x)=\phi(x)^T \hat{\Sigma} \phi(x)$ is a quadratic form of the feature embedding $\phi(x)$ with respect to the computed covariance $\hat{\Sigma}$, namely logit variance. Note that compared to a standard way of prediction during inference, the logit variance is additionally introduced to soften the prediction as temperature scaling.

After the standard model training, we compute the covariance $\hat{\Sigma}$ using the labeled dataset and estimate the optimal $\lambda$ by L-BFGS optimizer minimizing negative log-likelihood in the validation data set.

%% file: Appendix/D_FreeMatch.tex
\section*{Appendix D. Rationale in Class-wise Confidence Threshold}
\label{app: freematch}

By leveraging unlabeled data, semi-supervised learning has been highlighted to mitigate the requirement of substantial labeling costs and achieve comparable performance to a fully supervised setting with only a small amount of data \citep{yang2022survey}. Based on consistency regularization, the model acquires robustness to input perturbations, and the pseudo-labeling technique makes the decision boundary located in the low-density regions to separate the different class clusters.

FixMatch \citep{sohn2020fixmatch} is the most widely adopted algorithm in image classification and object detection. It uncovers the beneficial subset of the unlabeled data set for training using a fixed confidence threshold $\tau$ and two data augmentations of different strengths. Specifically, for every mini-batch of the unlabeled data set, it first computes the pseudo-label of weakly augmented input and only leaves the subset $S$ whose confidence is greater than or equal to $\tau$
\begin{equation*}
    \label{eq: subset}
    S=\{ x | \max(p_1,...p_C) \geq \tau  \quad \text{s.t.} \quad \forall c \in [C], \quad p_c = p\left(y=c|a(x),D\right) \}
\end{equation*}
where $a(\cdot)$ denotes the weak augmentation. Then, the strongly augmented input of the subset is used to train the model along with the mini-batch of the labeled data set.

As $\tau$ is generally set to a high value in FixMatch (for example, 0.95 in CIFAR-10/100), the convergence speed of FixMatch can be slow whenever the model struggles during training, especially in the early stage. Therefore, variants of FixMatch have been proposed to adaptively tune the confidence threshold $\tau$ online \citep{zhang2021flexmatch, wang2022freematch}. Notably, FreeMatch \citep{wang2022freematch} suggested to iteratively updating the global threshold $g^{(t)} \in \mathbb{R}$ and the local threshold $l^{(t)}=[l_{1}^{(t)}, ..., l_{C}^{(t)}] \in \mathbb{R}^{C}$ for every iteration $t$. Specifically, by initializing $g^{(0)}$ and every element of $l^{(0)}$ to $\frac{1}{C}$, the thresholds are updated using the model predictive confidence
\begin{subequations}
    \label{eq: global-local}
    \begin{align}
        g^{(t)} &= \eta \cdot g^{(t-1)} + (1-\eta) \cdot \frac{1}{\vert S \vert} \sum_{s=1}^{\vert S \vert} max(p_1,...,p_C) \\
        l_c^{(t)} &= \eta \cdot l_c^{(t-1)} + (1-\eta) \cdot \frac{1}{\vert S \vert} \sum_{s=1}^{\vert S \vert} p_c \quad \forall c \in [C]
    \end{align}
\end{subequations}
where $\eta$ is the coefficient of the exponential moving average. Then, the global threshold is modulated by the local threshold to compute the class-wise adaptive threshold.
\begin{equation}
    \label{eq: threshold}
    \tau_c^{(t)} = g^{(t)} \cdot l_c^{(t)} / max ( l_1^{(t)}, ..., l_C^{(t)} )
\end{equation}
We used the threshold to identify the instances of high aleatoric uncertainty by upper bounding the predictive confidence in the unlabeled data set.

%% file: Appendix/E_Settings.tex
\section*{Appendix E. Experimental Settings}
\label{app: settings}

We performed experiments in CIFAR-10/100 and SVHN, the commonly used image classification benchmarks in deep active learning. As a default data augmentation, we used random horizontal flip, random cropping followed by padding, and normalization.

The number of initially labeled data is set to 1000 in CIFAR-10, 5000 in CIFAR-100, and 1000 in SVHN, while the budget is set to 1000, 2500, and 1000, respectively. These values are inspected based on the insight that the optimal experimental settings to verify the effectiveness of active learning may differ across the data sets. We expect the initial accuracy to be over 40 and below 80 when trained by supervised learning. This is to obtain moderately reliable training signals and reflect the preference for considering both uncertainty and diversity. The budget size is adjusted to be either 50 or 100$\%$ of the number of the initially labeled data set, and the number of rounds is limited to 4.

Unless otherwise specified, Wide-ResNet-28-2 is used as a network architecture for the proxy model. When using supervised learning (i.e. DIFF), an SGD optimizer with weight decay 5e-4 is used for training 300 epochs with the learning rate scheduler that multiplies the initial learning rate of 0.1 by 0.5 at the 160-th, 240-th, and 280-th epoch. A batch size of 128 is employed. When using semi-supervised learning (i.e. SEMI, TBU), we mostly adopt the hyper-parameters from \cite{wang2022freematch}. To note, the total number of epochs is set to 200, where the number of iterations is somewhat reduced than the typical semi-supervised learning. This refers to \cite{coleman2019selection} for fast experiments with no significant difference in evaluating active learning algorithms. For fast convergence, a warm-up stage is added in CIFAR-100 and SVHN for 50 epochs of supervised learning with the labeled data where the learning rate is fixed to 0.1. 

Following the recommended settings proposed in \cite{munjal2022towards, ji2023randomness}, the target model is built upon ResNet-18 trained by an SGD optimizer with weight decay 5e-4 for 200 epochs. In CIFAR-10/100, the initial learning rate is set to 0.1 and multiplied by 0.1 at the 160-th epoch, while in SVHN, the learning rate is fixed to 0.01. A batch size of 128 is used. For an ablation study in \hyperref[app: results]{Appendix F}, VGG-16 is additionally considered as the target model. Keeping all other conditions identical to ResNet-18, the learning rate is fixed at 0.01 without scheduling.

%% file: Appendix/F_Results.tex
\section*{Appendix F. Additional Experimental Results}
\label{app: results}

In this section, we report (i) per-round accuracy in variations of data sets and target model architectures, (ii) robustness of TBU with respect to input perturbation, (iii) qualitative analysis on the LE and HA instances, and (iv) difficulty scheduling property of TBU which are not included in the main text due to space constraints.

\subsection*{F.1 Comparison of Per-round Accuracy with Variations}

\begin{figure}[h]
    \centering
    \begin{subfigure}[t]{\textwidth}
        \centering
        \includegraphics[width=\textwidth]{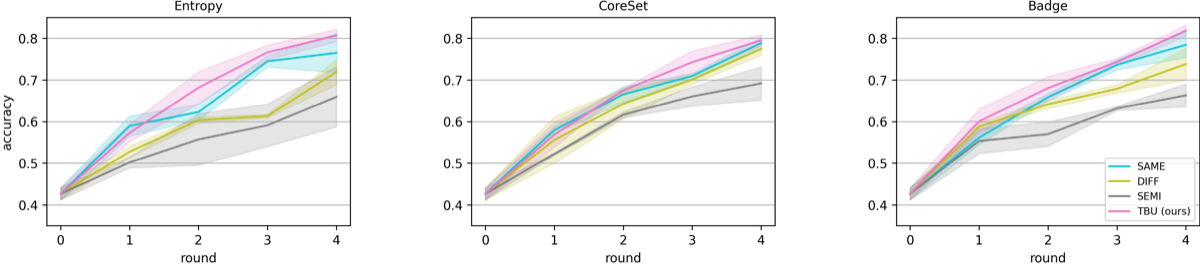}
        \caption{CIFAR-10 (I=1000, K=1000)}
    \end{subfigure}
    \vfill
    \begin{subfigure}[t]{\textwidth}
        \centering
        \includegraphics[width=\textwidth]{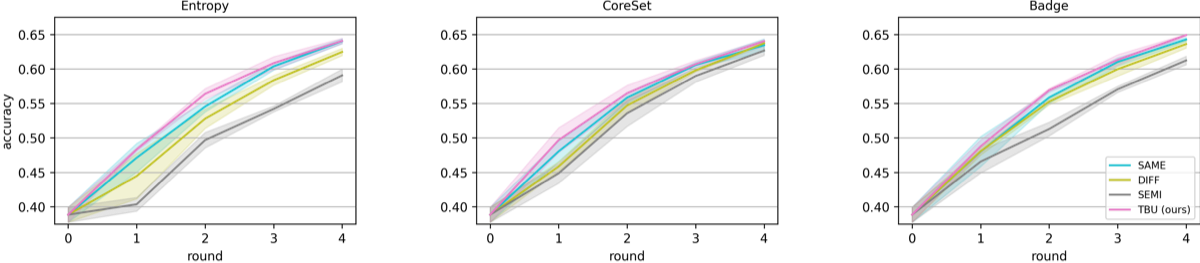}
        \caption{CIFAR-100 (I=5000, K=2500)}
    \end{subfigure}
    \vfill
    \begin{subfigure}[t]{\textwidth}
        \centering
        \includegraphics[width=\textwidth]{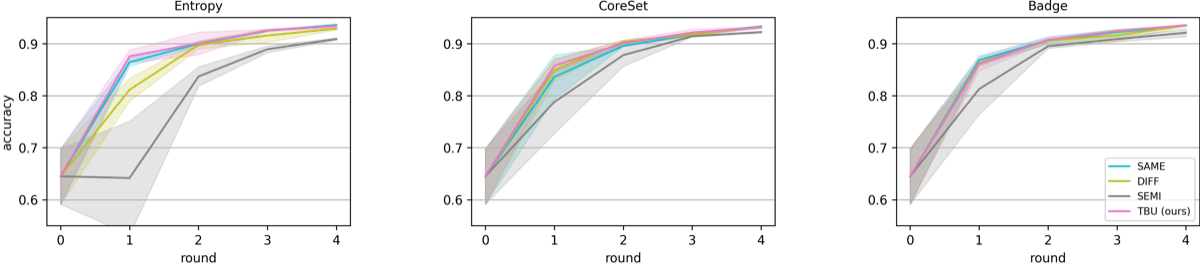}
        \caption{SVHN (I=1000, K=1000)}
    \end{subfigure}
    \caption{Per-round accuracy of the target model set to ResNet-18.}
    \label{fig: accuracy-res}
\end{figure}

\begin{table}[t]
    \centering
    \small
    \begin{tabular}{ |c|c|c|c|c|c|c| }
        \hline
        & & 1st & 2nd & 3rd & 4th \\
        \hline
        \multirow{4}{*}{Entropy} & SAME & \underline{0.471 $\pm$ 0.021} & \underline{0.545 $\pm$ 0.008} & \underline{0.604 $\pm$ 0.006} & \underline{0.640 $\pm$ 0.004} \\
        & DIFF & 0.445 $\pm$ 0.033 & 0.528 $\pm$ 0.014 & 0.584 $\pm$ 0.007 & 0.625 $\pm$ 0.005 \\
        & SEMI & 0.404 $\pm$ 0.010 & 0.497 $\pm$ 0.011 & 0.542 $\pm$ 0.004 & 0.591 $\pm$ 0.009  \\
        & TBU & \textbf{0.483 $\pm$ 0.002} & \textbf{0.564 $\pm$ 0.008} & \textbf{0.609 $\pm$ 0.009} & \textbf{0.641 $\pm$ 0.004} \\
        \hline
        \multirow{4}{*}{CoreSet} & SAME & \underline{0.481 $\pm$ 0.015} & \underline{0.558 $\pm$ 0.007} & \underline{0.606 $\pm$ 0.004} & 0.635 $\pm$ 0.009 \\
        & DIFF & 0.459 $\pm$ 0.007 & 0.547 $\pm$ 0.011 & 0.598 $\pm$ 0.002 & \underline{0.637 $\pm$ 0.001} \\
        & SEMI & 0.449 $\pm$ 0.014 & 0.536 $\pm$ 0.019 & 0.590 $\pm$ 0.009 & 0.627 $\pm$ 0.007 \\
        & TBU & \textbf{0.497 $\pm$ 0.018} & \textbf{0.565 $\pm$ 0.012} & \textbf{0.607 $\pm$ 0.005} & \textbf{0.640 $\pm$ 0.002} \\
        \hline
        \multirow{4}{*}{Badge} & SAME & 0.480 $\pm$ 0.022 & \underline{0.559 $\pm$ 0.008} & \underline{0.610 $\pm$ 0.006} & \underline{0.643 $\pm$ 0.003} \\
        & DIFF & \underline{0.481 $\pm$ 0.006} & 0.553 $\pm$ 0.005 & 0.600 $\pm$ 0.010 & 0.636 $\pm$ 0.006 \\
        & SEMI & 0.466 $\pm$ 0.017 & 0.513 $\pm$ 0.010 & 0.570 $\pm$ 0.005 & 0.612 $\pm$ 0.006 \\
        & TBU & \textbf{0.488 $\pm$ 0.009} & \textbf{0.569 $\pm$ 0.003} & \textbf{0.613 $\pm$ 0.008} & \textbf{0.649 $\pm$ 0.001} \\
        \hline
    \end{tabular}
    \caption{Per-round accuracy of the target model set to ResNet-18 in CIFAR-100.}
    \label{tab: accuracy-cifar100}
\end{table}

\begin{table}[t]
    \centering
    \small
    \begin{tabular}{ |c|c|c|c|c|c|c| }
        \hline
        & & 1st & 2nd & 3rd & 4th \\
        \hline
        \multirow{4}{*}{Entropy} & SAME & \underline{0.864 $\pm$ 0.008} & \underline{0.900 $\pm$ 0.005} & \underline{0.925 $\pm$ 0.001} & \textbf{0.935 $\pm$ 0.002} \\
        & DIFF & 0.811 $\pm$ 0.022 & 0.897 $\pm$ 0.008 & 0.915 $\pm$ 0.013 & 0.929 $\pm$ 0.002 \\
        & SEMI & 0.641 $\pm$ 0.109 & 0.837 $\pm$ 0.019 & 0.889 $\pm$ 0.006 & 0.909 $\pm$ 0.002 \\
        & TBU & \textbf{0.875 $\pm$ 0.012} & \textbf{0.900 $\pm$ 0.022} & \textbf{0.925 $\pm$ 0.002} & \underline{0.934 $\pm$ 0.003} \\
        \hline
        \multirow{4}{*}{CoreSet} & SAME & 0.836 $\pm$ 0.042 & 0.895 $\pm$ 0.001 & \underline{0.918 $\pm$ 0.003} & \textbf{0.933 $\pm$ 0.002} \\
        & DIFF & \underline{0.849 $\pm$ 0.020} & \textbf{0.904 $\pm$ 0.003} & 0.918 $\pm$ 0.002 & \underline{0.932 $\pm$ 0.002} \\
        & SEMI & 0.788 $\pm$ 0.061 & 0.878 $\pm$ 0.021 & 0.914 $\pm$ 0.001 & 0.922 $\pm$ 0.002 \\
        & TBU & \textbf{0.857 $\pm$ 0.013} & \underline{0.901 $\pm$ 0.006} & \textbf{0.921 $\pm$ 0.006} & 0.931 $\pm$ 0.001 \\
        \hline
        \multirow{4}{*}{Badge} & SAME & \textbf{0.868 $\pm$ 0.008} & \textbf{0.907 $\pm$ 0.007} & \underline{0.922 $\pm$ 0.005} & 0.934 $\pm$ 0.001 \\
        & DIFF & \underline{0.863 $\pm$ 0.006} & 0.906 $\pm$ 0.003 & 0.916 $\pm$ 0.012 & \underline{0.935 $\pm$ 0.001} \\
        & SEMI & 0.813 $\pm$ 0.049 & 0.895 $\pm$ 0.004 & 0.909 $\pm$ 0.005 & 0.921 $\pm$ 0.007 \\
        & TBU & 0.861 $\pm$ 0.013 & \underline{0.907 $\pm$ 0.004} & \textbf{0.924 $\pm$ 0.002} & \textbf{0.935 $\pm$ 0.002} \\
        \hline
    \end{tabular}
    \caption{Per-round accuracy of the target model set to ResNet-18 in SVHN. Given the same mean values, one with higher variance is marked as bold in preference to the maximal achievable performance.}
    \label{tab: accuracy-svhn}
\end{table}

\begin{figure}[h]
    \centering
    \begin{subfigure}[t]{\textwidth}
        \centering
        \includegraphics[width=\textwidth]{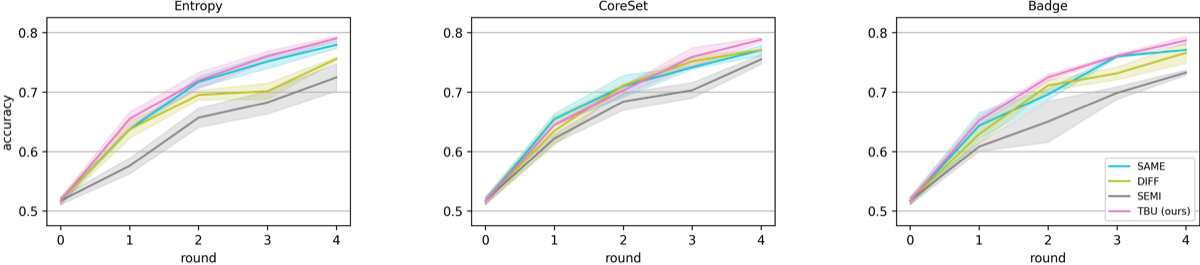}
        \caption{CIFAR-10 (I=1000, K=1000)}
    \end{subfigure}
    \vfill
    \begin{subfigure}[t]{\textwidth}
        \centering
        \includegraphics[width=\textwidth]{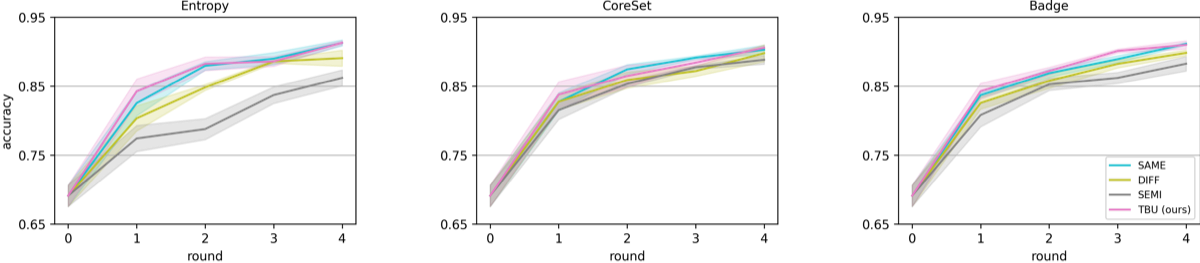}
        \caption{SVHN (I=1000, K=1000)}
    \end{subfigure}
    \caption{Per-round accuracy of the target model set to VGG-16.}
    \label{fig: accuracy-vgg}
\end{figure}

We compare the per-round accuracy with variations in data sets and target model architectures. In \autoref{fig: accuracy-res}, per-round accuracy in CIFAR-10/100 and SVHN is visualized. We also report the specific values in \autoref{tab: accuracy-cifar100} and \autoref{tab: accuracy-svhn} for CIFAR-100 and SVHN. Note that I and K indicate the number of initially labeled data and the budget for each selection round, respectively. Also, in \autoref{fig: accuracy-vgg}, we add an experimental result in CIFAR-10 and SVHN where the target model architecture is set to VGG-16. Note that solid lines indicate the average over three random seeds and the shaded region represents one standard deviation. TBU could outperform most of the baselines and verify its complementary effect on existing active learning algorithms.

\subsection*{F.2 Robustness of TBU in Corrupted Data Sets}

Due to the over-parameterized nature, deep neural networks are vulnerable to over-fitting, leading to wrong predictions while being over-confident. In order to tackle the over-fitting, many regularization techniques have emerged including dropout, batch normalization, and weight decay. However, it has been shown to be not enough in noisy labels \citep{song2022learning} and data set bias \citep{mehrabi2021survey} that are common in real-world data management. 

To equip robustness, finding an edge case becomes crucial to cover the under-explored regions of the underlying data distribution. Therefore, a model trained along with a desirable active learning algorithm is expected to be robust to input perturbations that implicitly lead to expanded data distribution.

\begin{figure}[h]
    \centering
    \begin{subfigure}[t]{\textwidth}
        \centering
        \includegraphics[width=\textwidth]{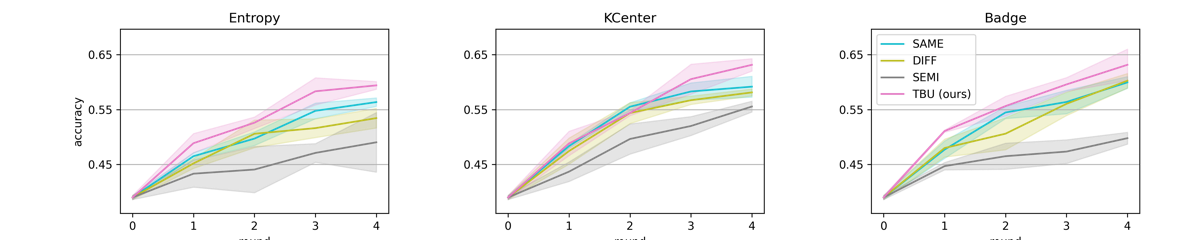}
        \caption{Gaussian blur}
    \end{subfigure}
    \vfill
    \begin{subfigure}[t]{\textwidth}
        \centering
        \includegraphics[width=\textwidth]{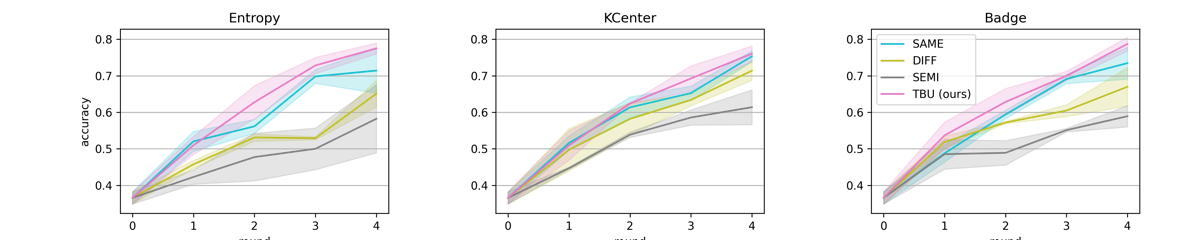}
        \caption{Brightness}
    \end{subfigure}
    \vfill
    \begin{subfigure}[t]{\textwidth}
        \centering
        \includegraphics[width=\textwidth]{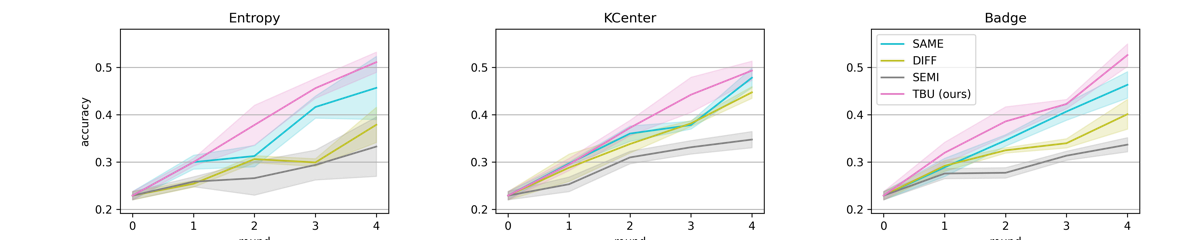}
        \caption{Contrast}
    \end{subfigure}
    \vfill
    \begin{subfigure}[t]{\textwidth}
        \centering
        \includegraphics[width=\textwidth]{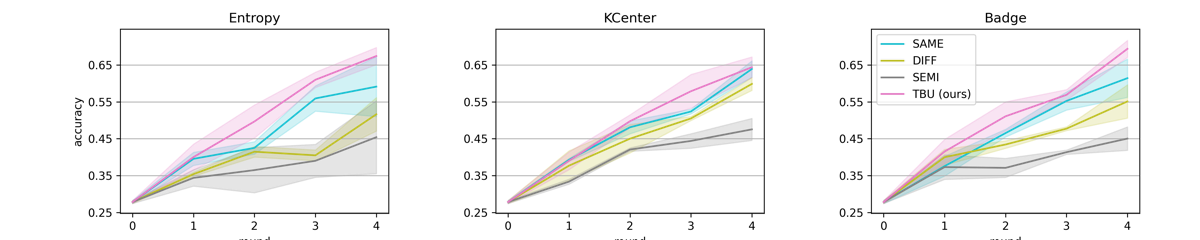}
        \caption{Fog}
    \end{subfigure}
    \vfill
    \begin{subfigure}[t]{\textwidth}
        \centering
        \includegraphics[width=\textwidth]{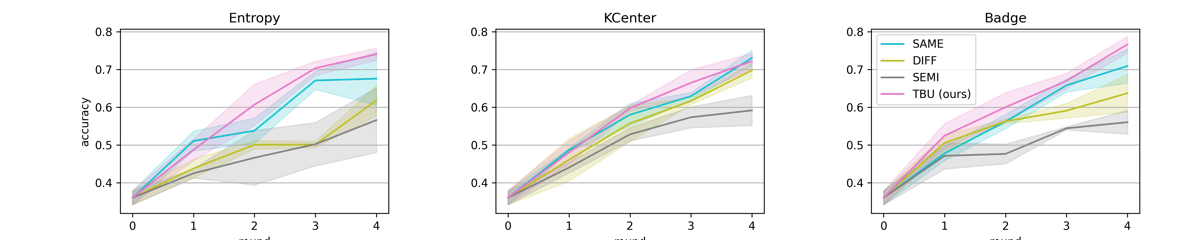}
        \caption{Saturate}
    \end{subfigure}
    \caption{Per-round accuracy of the target model set to ResNet-18 in CIFAR-10-C.}
    \label{fig: robustness-cifar10}
\end{figure}

\begin{figure}[h]
    \centering
    \begin{subfigure}[t]{\textwidth}
        \centering
        \includegraphics[width=\textwidth]{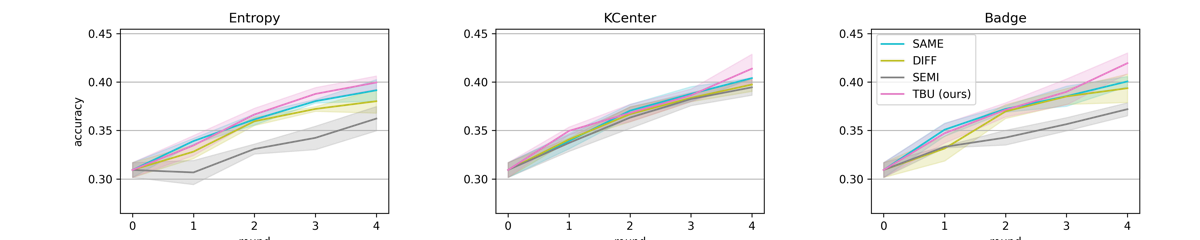}
        \caption{Gaussian blur}
    \end{subfigure}
    \vfill
    \begin{subfigure}[t]{\textwidth}
        \centering
        \includegraphics[width=\textwidth]{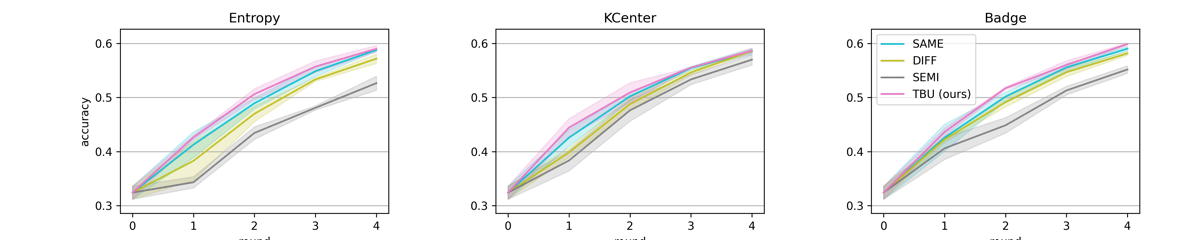}
        \caption{Brightness}
    \end{subfigure}
    \vfill
    \begin{subfigure}[t]{\textwidth}
        \centering
        \includegraphics[width=\textwidth]{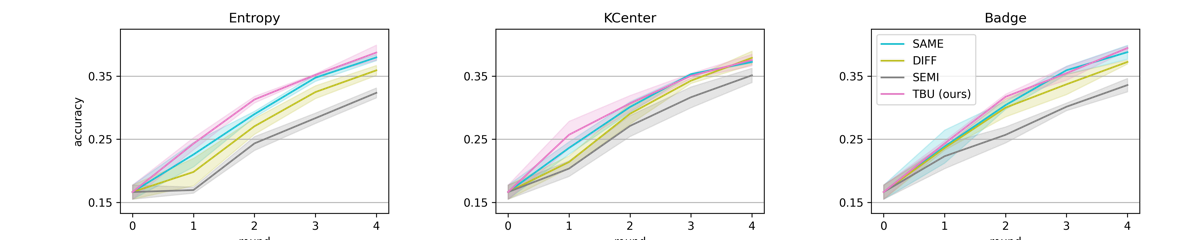}
        \caption{Contrast}
    \end{subfigure}
    \vfill
    \begin{subfigure}[t]{\textwidth}
        \centering
        \includegraphics[width=\textwidth]{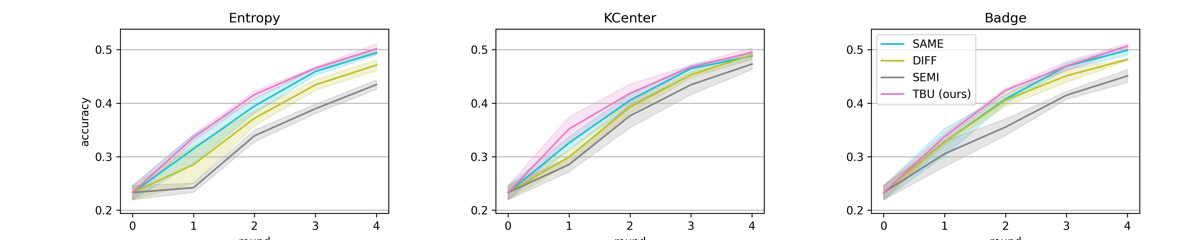}
        \caption{Fog}
    \end{subfigure}
    \vfill
    \begin{subfigure}[t]{\textwidth}
        \centering
        \includegraphics[width=\textwidth]{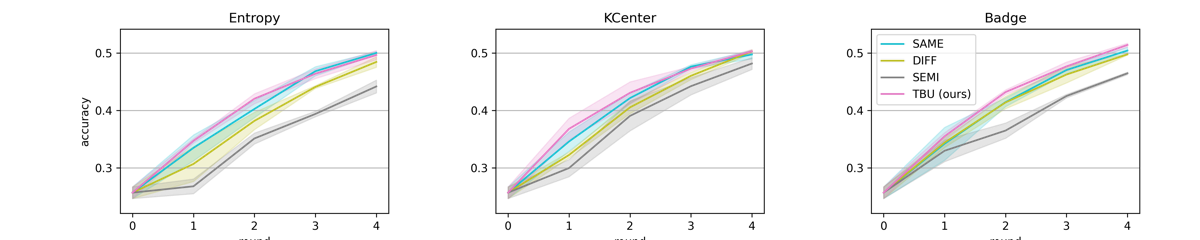}
        \caption{Saturate}
    \end{subfigure}
    \caption{Per-round accuracy of the target model set to ResNet-18 in CIFAR-100-C.}
    \label{fig: robustness-cifar100}
\end{figure}

CIFAR-10/100-C \citep{hendrycks2019benchmarking} is the frequently used data set to verify the robustness of model prediction to input perturbation. Among various options, we chose five different corruption types (gaussian blur, brightness, contrast, fog, and saturate) for generality. In \autoref{fig: robustness-cifar10} and \autoref{fig: robustness-cifar100}, per-round accuracy evaluated in CIFAR-10/100-C is reported, showing the superiority of TBU.

\subsection*{F.3. Qualitative Analysis on LE and HA instances}

\begin{figure}[h]
     \centering
     \begin{subfigure}[c]{0.45\textwidth}
         \centering
         \includegraphics[width=\textwidth]{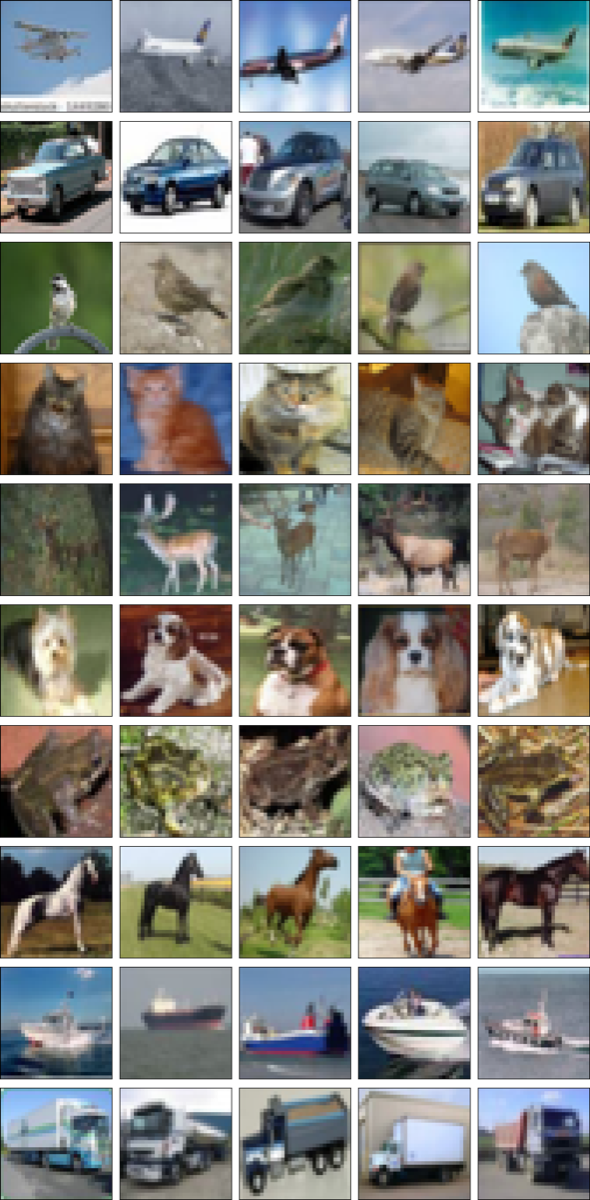}
         \caption{LE instances}
         \label{fig: easy_examples}
     \end{subfigure}
     \hfill
     \begin{subfigure}[c]{0.45\textwidth}
         \centering
         \includegraphics[width=\textwidth]{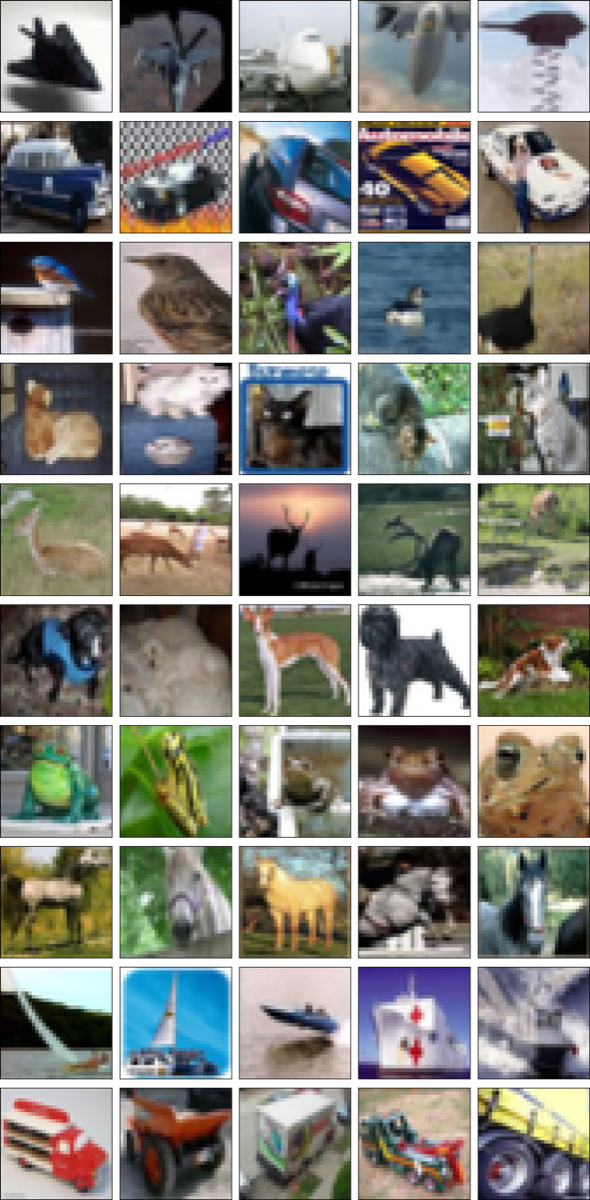}
         \caption{HA instances}
         \label{fig: hard_examples}
     \end{subfigure}
    \caption{Example images of the LE and HA instances from CIFAR-10.}
    \label{fig: easy_and_hard_examples}
\end{figure}

We attach five randomly chosen images from the LE and HA instances in \autoref{fig: easy_and_hard_examples} where each row comprises images of the same ground-truth class label. LE instances show stronger semantic similarity in their imagery, while HA instances incorporate unusual patterns like fragmented object components. In short, LE instances can be easily substituted by others, and HA instances can lead to incorrect inductive bias in the model training. Particularly, in the early rounds of the active learning cycle, the labeled data is scarce and more instances in the unlabeled data tend to be assigned to HA instances. In contrast, in the later rounds, LE instances would become dominant in numbers as the model prediction gets more accurate and calibrated.

\subsection*{F.4 Difficulty Scheduling Property of TBU}

\begin{table}[t]
    \centering
    \scriptsize
    \begin{tabular}{ |c|c|c|c|c|c|c|c| }
        \hline
        \multirow{2}{*}{Data} & \multirow{2}{*}{Round} & \multicolumn{2}{|c|}{Entropy} & \multicolumn{2}{|c|}{CoreSet} & \multicolumn{2}{|c|}{Badge} \\
        \cline{3-8}
        & & LE & HA & LE & HA & LE & HA \\
        \hline
        \multirow{4}{*}{CIFAR-10} & 1st & 8.0 $\pm$ 1.2 & \textbf{34.3 $\pm$ 6.9} & 8.0 $\pm$ 1.2 & \textbf{34.3 $\pm$ 6.9} & 8.0 $\pm$ 1.2 & \textbf{34.3 $\pm$ 6.9} \\
        & 2nd & 10.4 $\pm$ 1.2 & 12.9 $\pm$ 2.2 & 10.0 $\pm$ 0.6 & 9.3 $\pm$ 0.6 & 10.2 $\pm$ 0.5 & 13.5 $\pm$ 1.3 \\ 
        & 3rd & 13.5 $\pm$ 0.8 & 10.5 $\pm$ 1.2 & 12.0 $\pm$ 0.3 & 9.1 $\pm$ 0.7 & 12.3 $\pm$ 0.4 & 10.3 $\pm$ 1.2 \\ 
        & 4th & \textbf{16.0 $\pm$ 0.9} & 9.1 $\pm$ 1.2 & \textbf{13.1 $\pm$ 1.1} & 8.8 $\pm$ 0.8 & \textbf{13.2 $\pm$ 0.6} & 8.3 $\pm$ 1.7 \\ 
        \hline
        \multirow{4}{*}{CIFAR-100} & 1st & 5.2 $\pm$ 0.3 & 7.0 $\pm$ 0.4 & 5.2 $\pm$ 0.3 & 7.0 $\pm$ 0.4 & 5.2 $\pm$ 0.3 & 7.0 $\pm$ 0.4 \\
        & 2nd & 8.1 $\pm$ 0.1 & \textbf{7.1 $\pm$ 0.2} & 7.8 $\pm$ 0.1 & \textbf{7.5 $\pm$ 0.3} & 7.9 $\pm$ 0.2 & 7.0 $\pm$ 0.5 \\ 
        & 3rd & 11.8 $\pm$ 0.4 & 7.0 $\pm$ 0.6 & 10.8 $\pm$ 0.6 & 7.1 $\pm$ 0.7 & 10.9 $\pm$ 0.3 & \textbf{7.1 $\pm$ 0.6} \\ 
        & 4th & \textbf{15.5 $\pm$ 0.4} & 5.4 $\pm$ 0.6 & \textbf{12.9 $\pm$ 0.2} & 6.5 $\pm$ 0.1 & \textbf{14.3 $\pm$ 0.5} & 5.9 $\pm$ 0.2 \\ 
        \hline
        \multirow{4}{*}{SVHN} & 1st & 17.7 $\pm$ 1.1 & \textbf{18.0 $\pm$ 2.3} & 17.7 $\pm$ 1.1 & \textbf{18.0 $\pm$ 2.3} & 17.7 $\pm$ 1.1 & \textbf{18.0 $\pm$ 2.3} \\
        & 2nd & 28.0 $\pm$ 0.6 & 11.6 $\pm$ 0.7 & 22.8 $\pm$ 1.6 & 9.6 $\pm$ 1.1 & 26.0 $\pm$ 1.2 & 9.2 $\pm$ 2.0 \\ 
        & 3rd & 36.2 $\pm$ 0.6 & 4.2 $\pm$ 0.4 & 28.7 $\pm$ 1.5 & 5.1 $\pm$ 0.2 & 31.3 $\pm$ 1.5 & 5.1 $\pm$ 0.8 \\ 
        & 4th & \textbf{40.7 $\pm$ 0.6} & 2.1 $\pm$ 0.1 & \textbf{31.9 $\pm$ 0.8} & 2.7 $\pm$ 0.7 & \textbf{32.5 $\pm$ 1.4} & 2.3 $\pm$ 0.1 \\ 
        \hline
    \end{tabular}
    \caption{Per-round percentage of the LE and HA instances in the unlabeled data.}
    \label{tab: e_h_schedule}
\end{table}

Trivially evident data instances guide the inductive bias about the underlying data distribution. In contrast, edge case data can contribute to model calibration by alleviating overconfidence and robustifying decision boundaries. According to theoretical analysis of training dynamics \citep{oymak2020toward, wang2022deep}, easy instances are particularly effective in the early phase of model training, which gradually hands over the attention to hard instances.

Regarding the LE instances as the easy and the HA instances as the hard, TBU was able to schedule the difficulty of subsets along selection rounds adaptively. In \autoref{tab: e_h_schedule}, the percentage of LE and HA instances in the unlabeled data are reported in CIFAR-10/100 and SVHN. Commonly, as the active learning cycle proceeds, LE instances increase and HA instances decrease. As a result, the remaining instances, neither LE nor HA, can gradually accommodate the instances regarded as HA instances in previous rounds.